\pdfoutput=1

\documentclass[11pt]{article}

\usepackage{acl}
\usepackage{times}
\usepackage{latexsym}
\usepackage[T1]{fontenc}
\usepackage[utf8]{inputenc}
\usepackage{microtype}
\usepackage{inconsolata}
\usepackage{graphicx}
\usepackage{hyperref}       
\usepackage{url}            
\usepackage{booktabs}       
\usepackage{amsfonts}       
\usepackage{nicefrac}       
\usepackage{enumitem}
\usepackage{multicol}
\usepackage{multirow}
\usepackage{amsmath}
\usepackage{amssymb}
\usepackage{bbding}
\usepackage{subfigure}
\usepackage{colortbl}
\usepackage{pifont}
\usepackage{makecell}
\usepackage{mathrsfs}
\usepackage{bm}
\usepackage{bbm}
\usepackage{dsfont}
\usepackage{pgf}
\usepackage{fontawesome}
\usepackage{wrapfig}
\usepackage{tcolorbox}
\usepackage{array}
\usepackage{longtable} 
\usepackage{caption}
\usepackage{siunitx}
\newcommand{\ourmethod}{\textsc{Cat}}
\newcommand{\instruct}{\textsc{CaT}-Instruct}
\newcommand{\baseinstruct}{\textsc{Base-Instruct}}
\newcommand{\datamethod}{\textsc{CaT-Generator}}
\newcommand{\ourmodel}{SWE-Compressor}

\title{Context as a Tool: Context Management for Long-Horizon SWE-Agents}

\author{
  Shukai Liu\textsuperscript{\rm 1},
  {\bf Jian Yang}\textsuperscript{\rm 1\thanks{Corresponding author.}},
  {\bf Bo Jiang}\textsuperscript{\rm 1*},
  {\bf Yizhi Li}\textsuperscript{\rm 2},
  {\bf Jinyang Guo}\textsuperscript{\rm 1}, \\
  {\bf Xianglong Liu}\textsuperscript{\rm 1},
  {\bf Bryan Dai}\textsuperscript{\rm 3}\\
   \textsuperscript{\rm 1}Beihang University;
   \textsuperscript{\rm 2}Manchester;
   \textsuperscript{\rm 3}Ubiquant;
   \\
   \texttt{\{skliu,jiayang\}@buaa.edu.cn} \\
}

\begin{document}
\maketitle
\begin{abstract}

Agents based on large language models have recently shown strong potential on real-world software engineering (SWE) tasks that require long-horizon interaction with repository-scale codebases. However, most existing agents rely on append-only context maintenance or passively triggered compression heuristics, which often lead to context explosion, semantic drift, and degraded reasoning in long-running interactions.
We propose \ourmethod{}, a new context management paradigm that elevates context maintenance to a callable tool integrated into the decision-making process of agents. \ourmethod{} formalizes a structured context workspace consisting of stable task semantics, condensed long-term memory, and high-fidelity short-term interactions, and enables agents to proactively compress historical trajectories into actionable summaries at appropriate milestones.
To support context management for SWE-agents, we propose a trajectory-level supervision framework, \datamethod{}, based on an offline data construction pipeline that injects context-management actions into complete interaction trajectories.
Using this framework, we train a context-aware model, \ourmodel{}. Experiments on SWE-Bench-Verified demonstrate that \ourmodel{} reaches a 57.6\% solved rate and significantly outperforms ReAct-based agents and static compression baselines, while maintaining stable and scalable long-horizon reasoning under a bounded context budget.

\end{abstract}

\section{Introduction}
\label{sec:introduction}
Large language models (LLMs)~\cite{yang2025qwen3,touvron2023llama,anthropic2025claude_sonnet45,achiam2023gpt} have achieved remarkable progress on tasks such as code generation and bug fixing~\cite{chen2021evaluating,austin2021program,liu2024mdeval,mceval}. As research attention increasingly shifts toward real-world software engineering (SWE) scenarios, maintaining stable and effective reasoning in complex and long-horizon interactive tasks has emerged as a primary challenge in agent research. Most SWE tasks (e.g., repository-level issue resolution~\cite{jimenez2023swe}) often require LLMs to continuously interpret environment feedback, execute actions, and revise strategies over hundreds of interaction rounds, which poses greater demands on context management.

\begin{figure}[t]
\centering
\includegraphics[width=1\linewidth]{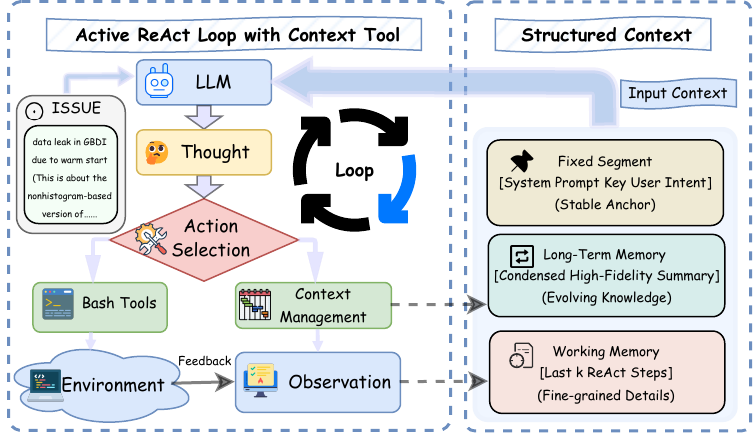}
\caption{Overview of \ourmethod{} with a structured context workspace for long-horizon reasoning.}
\vspace{-20pt}
\label{fig:intro}
\end{figure}

Most existing code agents based on paradigms such as ReAct~\cite{yao2022react} adopt an append-only context maintenance strategy, where all past interactions are continuously concatenated into the messages. While this method may suffice for short-horizon tasks, it often leads to rapid context expansion in long-horizon scenarios, resulting in information redundancy, semantic drift, and even reasoning collapse. To mitigate these challenges, prior work~\cite{jiang2023llmlingua,shinn2023reflexion,wang2025recursively,packer2023memgpt} has explored context compression, summarization, or multi-level memory mechanisms to constrain context size. Nevertheless, previous approaches treat context management as a passively triggered heuristic mechanism, so they lack the flexibility to adapt compression timing and content across different task phases, limiting their adaptability and scalability in complex environments.
In this paper, we propose a new context management paradigm, \ourmethod{}.
We argue that in long-horizon interactive tasks, context management should be internalized as a model capability rather than enforced through external constraints.
Motivated by this perspective, \ourmethod{} treats context management as a callable and plannable tool, on par with environment-interaction tools such as code editing and command execution, which integrates context maintenance into the action process as an active and learnable component.

In~\autoref{fig:intro}, \ourmethod{} organizes the context into a structured workspace consisting of stable task-semantic anchors, an evolvable long-term memory, and a short-term working memory. It further enables the agent to proactively trigger context folding at stage boundaries, compressing redundant histories into high-fidelity and actionable long-term memory representations.
Building on this design, we introduce a trajectory-level supervision framework, \datamethod{}, which injects context-management behaviors into complete interaction trajectories via offline reconstruction. Using \datamethod{}, we train a context-aware model, \ourmodel{}, allowing it to learn when to compress context, how to generate effective summaries, and how to reuse compressed representations during subsequent reasoning.

Experimental results on SWE-Bench show that \ourmethod{} substantially outperforms ReAct without context management and baselines with static compression strategies under the same model scale and interaction budget, demonstrating the strong context scalability in long-horizon interaction settings. The contribution of his paper is summarized as:
\begin{itemize}
    \item We propose \ourmethod{}, a new context management paradigm that internalizes context maintenance as a learnable, tool-based capability for long-horizon interactive reasoning.
    \item We design a structured context workspace with proactive context folding, enabling effective memory construction and sustained reasoning under bounded context budgets.
    \item We introduce \datamethod{}, a trajectory-level supervision framework for learning context-management behaviors, and train a context-aware model, \ourmodel{}, that effectively compresses and reuses context during extended interactions.
\end{itemize}

\begin{figure}[t]
\centering
\includegraphics[width=1\linewidth]{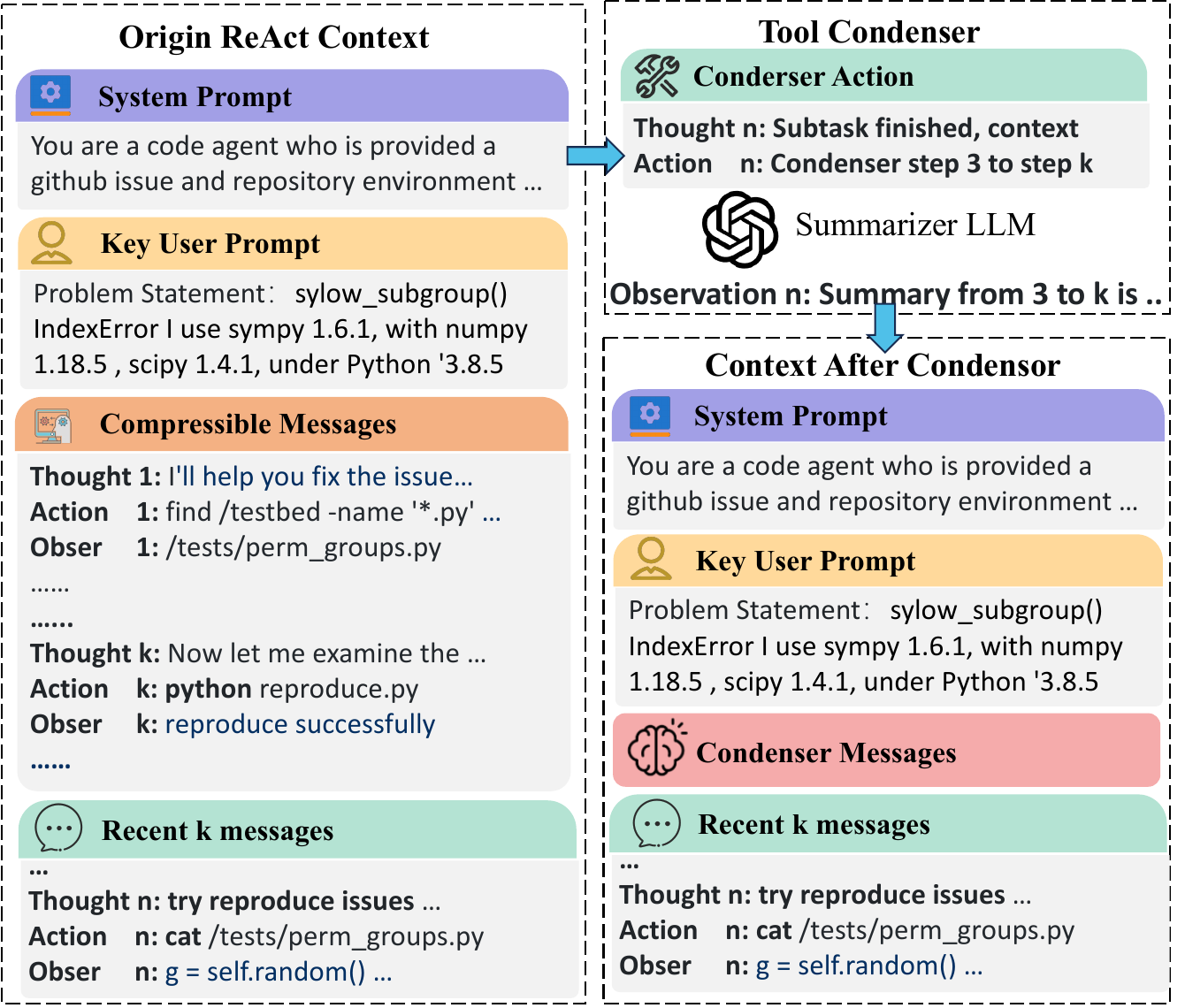}
\caption{Example of structured context condensation in \ourmethod{} during long-horizon tasks.}
\vspace{-15pt}
\label{fig:example}
\end{figure}

\section{\ourmodel{}}
\begin{figure*}[h]
\begin{center}
    \includegraphics[width=0.9\textwidth]{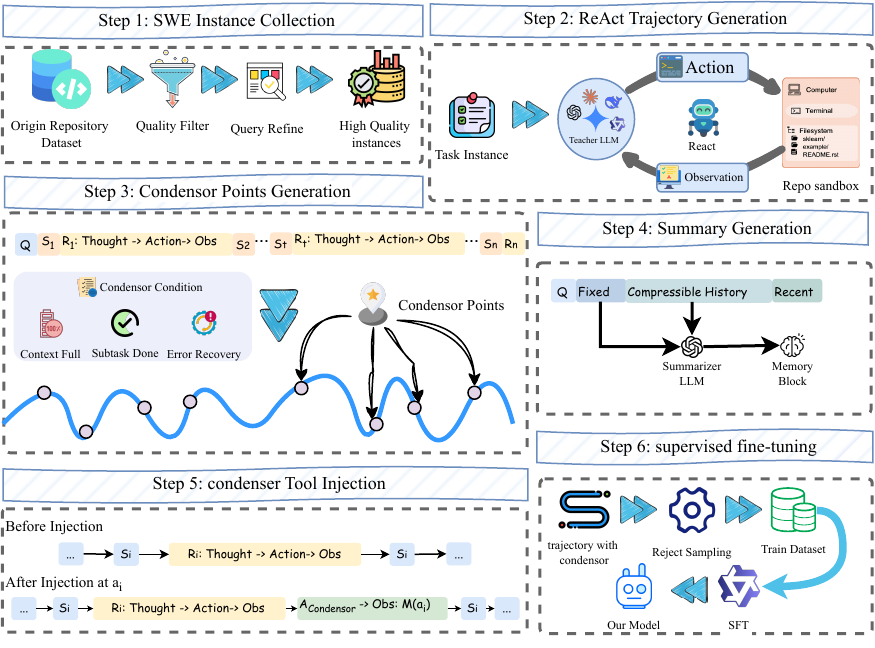}
    \vspace{-5pt}
    \caption{Overview of the data construction and training pipeline for \ourmethod{}. The process includes SWE instance collection, base ReAct trajectory generation, condenser point identification, structured summary generation, tool-based context injection, and supervised fine-tuning with rejection sampling.}
    \label{fig:sft}
    \vspace{-15pt}
\end{center}
\end{figure*}

Unlike prior agents treating context management as a passive heuristic or post-processing step, \ourmethod{} elevates it to a callable and plannable capability on par with environment-interaction tools, making context maintenance an active and learnable component of the agent’s decision-making process. 
Specifically, we formalize a structured context workspace, introduce tool-based context management with segmented compression for ReAct, and present a trajectory-level supervision framework~(\autoref{fig:sft}) with an offline data pipeline that enables effective context compression and reuse in subsequent reasoning and decision-making.

\subsection{Structured Context Workspace}
When performing long-horizon ReAct reasoning in complex environments, agent performance largely depends on how its working context is organized. Motivated by this observation, \ourmethod{} models context as a controllable and dynamically updated cognitive workspace composed of three functional segments, as illustrated in~\autoref{fig:example}. The first is a fixed segment that preserves the system prompt and key user intent. The second is a long-term memory segment that stores a condensed, high-fidelity summary of historical trajectories. The third is a high-fidelity working memory segment that retains the most recent $k$ ReAct interaction steps. This design provides a stable semantic anchor for the task while preserving fine-grained information from recent environment feedback, thereby supporting precise contextualized actions.
Formally, the working context at step $t$ is represented as
\begin{MiddleEquation}
\begin{align}
    C(t)=\left(Q,\, M(t),\, I^{(k)}(t)\right)
\end{align}
\end{MiddleEquation}where $Q$ denotes the non-compressible component consisting of the system prompt and key user objectives, $M(t)$ represents the high-fidelity summary of historical trajectories, and $I^{(k)}(t)$ denotes the complete records of the most recent $k$ ReAct interactions. At initialization, $C(1)=(Q,\emptyset,\emptyset)$. As reasoning progresses, the agent continuously updates recent interactions and long-term summaries, and adjusts the content and structure of $M(t)$ through the context-management tool when needed. In this way, stable goals, condensed knowledge, and fine-grained working memory are coordinated within a unified framework, mitigating both semantic drift and uncontrolled context expansion.

\subsection{Context Management as a First-Class Tool}
The key innovation of \ourmethod{} is to explicitly model context management as a callable tool operation and to place it at the same decision level as environment-interaction tools such as file editing or command execution. Consequently, when generating a response at step $t$, the agent evaluates whether to invoke the context-management tool in the same manner as it selects other external actions. In practice, the agent tends to proactively trigger context management when a subtask has been completed and requires a stage-wise summary, when the trajectory has grown sufficiently large that historical compression is necessary to maintain operational efficiency, or when subsequent reasoning benefits more from a concise, structured summary than from verbose raw logs. Through this design, context management is transformed from a post hoc procedure into a self-regulating reasoning step, becoming an integral part of the agent’s strategy rather than an external constraint.

\subsection{Structured Memory Generation}
For the compressible historical segment, \ourmethod{} applies structured summarization to condense information accumulated during long-horizon reasoning into a compact and actionable long-term memory. The goal is to reduce context length while preserving information that remains causally relevant for subsequent decisions.
The resulting long-term memory summarizes key aspects of task progress, including intermediate goals, adopted strategies and their outcomes, salient environment feedback, and persistent constraints that continue to shape future reasoning. 
After summarization, the agent reconstructs its working context by combining the fixed task semantics, the condensed long-term memory, and the most recent high-fidelity interaction steps, enabling stable and consistent reasoning under a bounded context budget.

\subsection{Supervised Trajectory Construction}

\paragraph{Data Generation Pipeline}
To internalize toolized context management capability in LLM, we propose \datamethod{}, a two-stage retrospective trajectory construction pipeline, as illustrated in~\autoref{fig:sft}. \datamethod{} first generates complete base ReAct reasoning trajectories without introducing any context compression operations, to maximize the naturalness and completeness of task-solving behaviors. Then, these trajectories are minimally and structurally reconstructed: context management tool invocations are injected at appropriate steps without altering the original sequence of environment interactions, yielding SFT data that are consistent with the reasoning paradigm of \ourmethod{}.

\paragraph{Phase I: Base ReAct Trajectory Generation.}
Given a task instance, we first deploy a standard ReAct agent to execute a complete interaction process in a controlled environment, while explicitly disabling the context management tool and allowing only environment-related actions (e.g., code editing, command execution, or information retrieval). The resulting raw trajectory is denoted as
\begin{SmallEquation}
\begin{align}
\mathcal{T}_{\text{base}} = \{(C_{\text{base}}(1), R_{\text{base}}(1)), \ldots, (C_{\text{base}}(T), R_{\text{base}}(T))\}
\end{align}
\end{SmallEquation}where $R_{\text{base}}(t)$ follows the standard ReAct response structure, consisting of Thought and Action. The objective of this stage is to preserve as much intermediate reasoning, failure patterns, and environment feedback as possible, thereby providing sufficient causal evidence for learning context compression decisions in later stages.

\paragraph{Phase II: Trajectory Refactoring by injecting Compression Operation.}

\begin{table}[t]
    \centering
    \footnotesize
    \setlength{\tabcolsep}{8pt}
    \begin{tabular}{l r@{}}
    \toprule
    \textbf{Statistic} & \textbf{\instruct{}} \\
    \midrule
    \multicolumn{2}{l}{\textit{Trajectory length (steps)}} \\
    \quad Average & 87.4 \\
    \quad Median  & 77.5 \\
    \quad Max     & 500 \\
    \midrule
    \multicolumn{2}{l}{\textit{Context size (tokens)}} \\
    \quad Avg tokens per step & 13,044 \\
    \quad Median tokens per step  & 10,875 \\
    \quad Max tokens per step     & 65,536 \\
    \midrule
    \multicolumn{2}{l}{\textit{Context-mgmt actions / traj}} \\
    \quad Average & 4.22 \\
    \quad Median  & 4.00 \\
    \quad Max     & 26 \\
    \midrule
    \multicolumn{2}{l}{\textit{Compression}} \\
    \quad Avg tokens before & 15,585 \\
    \quad Avg tokens after  & 4,676 \\
    \quad Avg ratio (\%)    & 30 \\
    \bottomrule
    \end{tabular}
    \caption{Statistics of SFT data for \instruct{}.}
    \vspace{-15pt}
    \label{tab:training_data_stats}
\end{table}

We perform an offline reconstruction of the base trajectory $\mathcal{T}_{\text{base}}$ to obtain an augmented trajectory $\mathcal{T}_{\text{retro}}$ that includes context management tool invocations. This stage is to transform context compression from an implicit side effect during generation into an explicitly modeled, controllably injected tool operation. 

\paragraph{(1) Condensor Position Generation.}
We begin by conducting a structured analysis of the base trajectory to identify a set of candidate time steps suitable for inserting context management tool calls, denoted as $\mathcal{A}=\{a_1,\ldots,a_m\}$. The selection of insertion position jointly considers multiple signals, including: (i) context expansion signals, such as sustained growth in context length or decreasing token utilization efficiency; (ii) structural boundary signals, such as subtask completion, strategy switching, or intermediate milestones; and (iii) error-correction signals, where new feasible directions emerge after repeated failures. We treat these signals as heuristic triggers rather than fixed thresholds, aligning insertion points with natural moments for stage-wise summarization in long-horizon reasoning and improving the learnability of compression behaviors.

\paragraph{(2) Segmented Context Construction and Compression Input Preparation.}
For each candidate insertion position $a_i$, we construct a segmented representation of the context at that step by partitioning the visible context into a fixed segment $Q$, a recent high-fidelity working memory segment $I^{(k)}(a_i)$, and a compressible historical trajectory segment. The fixed segment and the recent trajectory segment are preserved verbatim, while the historical segment is provided as input to model for long-term memory summarization. This procedure enforces the structured context workspace constraints of \ourmethod{} during the offline stage.

\paragraph{(3) Long-Term Memory Block Generation.}
Based on the segmented context, we invoke a high-capacity language model to generate a structured long-term memory block $M(a_i)$. The summarizer uses the same backbone as the reasoning model (\ourmodel{}), keeping summarization aligned with the agent’s internal reasoning style. This memory block aims to faithfully summarize critical information from the compressible history, including completed subtasks, attempted strategies and their outcomes, important environment state changes, facts that continue to constrain subsequent decisions, and key information that remains useful for future steps. The generated $M(a_i)$ serves as the Observation of a context management tool invocation and is written into the long-term memory segment for subsequent reasoning.

\paragraph{(4) Trajectory Stitching and Minimal-Intrusion Compression Injection.}
Finally, we adopt a minimal-intrusion trajectory stitching strategy to inject context management behaviors into the original ReAct trajectory. Specifically, for each insertion point $a_i$, we explicitly insert a context management tool invocation at that step as an independent Action, whose corresponding Observation is the generated long-term memory block $M(a_i)$.
\begin{MiddleEquation}
$
\mathcal{T}=\{(C(1),R(1)),(C(2),R(2)),\ldots,(C(T),R(T))\}
$
\end{MiddleEquation}
where $T$ denotes the total number of steps required to complete the task. Each trajectory fully captures the process from the initial task specification, through multiple rounds of environment interaction and stage-wise context compression, to final task completion. The advantage of trajectory-level supervision lies in preserving the temporal continuity of context evolution, allowing the model not only to observe the immediate summary produced by a context-management tool invocation but also to learn how that summary influences reasoning and decision-making across subsequent steps. Moreover, it enables the model to learn cross-step strategic judgments, such as how different invocation timings affect downstream efficiency and stability, thereby better reflecting the decision structure of real long-horizon tasks.
\paragraph{(5) Rejection Sampling Fine-Tuning.}
To construct high-quality trajectory-level SFT data, we apply a rejection sampling strategy during data curation to filter interaction trajectories using trajectory-level and step-level criteria. At the trajectory level, samples that fail to complete the task or enter unrecoverable error states are discarded. At the step level, we further remove trajectories exhibiting unreasonable context-management behaviors, such as excessively frequent tool invocations with minimal information gain, severe semantic drift, or internal state inconsistencies. 
The resulting curated set of trajectories constitutes our SFT dataset (\instruct{}). \autoref{tab:training_data_stats} summarizes key statistics of \instruct{}, including trajectory length, context size, and the frequency and effectiveness of context-management actions.
Using \instruct{} for supervised fine-tuning, we obtain \ourmodel{}, which internalizes context management as a learned model capability.

\definecolor{TableHeaderColor}{HTML}{D1D5DB}    
\definecolor{HeaderColor}{HTML}{E5E7EB}        
\definecolor{React100BColor}{HTML}{E3F2FD}      
\definecolor{ReactColor}{HTML}{FFF3E0}          
\definecolor{SummaryColor}{HTML}{F9FBE7}        
\definecolor{FoldingColor}{HTML}{F3E5F5}        

\begin{table*}[h]
    \centering
    \small
    \renewcommand{\arraystretch}{1.0}
    \setlength{\tabcolsep}{12pt}
    \resizebox{0.9\textwidth}{!}{%
    \begin{tabular}{lc|cc}
    \toprule
     \multicolumn{2}{c|}{\textbf{Method}}  & \multicolumn{2}{c}{\textbf{SWE-Bench Verified}} \\
     \midrule
    \textbf{Model} & \textbf{Model Size} & \textbf{Scaffold} & \textbf{Pass@1} \\
    \rowcolor{HeaderColor}\multicolumn{4}{c}{\textbf{ReAct Agent with 100B+ LLM}}  \\
    \rowcolor{React100BColor}
    GPT-5.1~\cite{openai2025gpt51}   & \faLock{} & OpenHands & 76.3   \\
    \rowcolor{React100BColor}
    GPT-4o~\cite{gpt4} & \faLock{} & Agentless & 38.8   \\
    \rowcolor{React100BColor}
    Claude-3.5-Sonnet~\cite{anthropic2024claude35sonnet} & \faLock{} & OpenHands & 53.0   \\
    \rowcolor{React100BColor}
    Claude-4.5-Sonnet~\cite{anthropic2025claude_sonnet45} & \faLock{} & OpenHands & 77.2   \\
    \rowcolor{React100BColor}
    Gemini-2.5-Pro~\cite{google2025gemini25pro} & \faLock{} & OpenHands & 59.6   \\
    \rowcolor{React100BColor}
    Gemini-3-Pro~\cite{deepmind2025gemini3pro} & \faLock{} & OpenHands & 76.2   \\
    \rowcolor{HeaderColor}\multicolumn{4}{c}{\textbf{ReAct Agent}} \\
    \rowcolor{ReactColor}
    R2E-Gym-32B~\cite{jain2025r2e} & 32B & OpenHands & 34.4   \\
    \rowcolor{ReactColor}
    SWE-Gym-32B~\cite{pan2024training} & 32B & OpenHands & 20.6   \\
    \rowcolor{ReactColor}
    SWE-agent-LM-32B~\cite{yang2025swe} & 32B & SWE-agent & 40.2   \\
    \rowcolor{ReactColor}
    DeepSWE-32B-Preview~\cite{deepswe2025} & 32B & OpenHands & 42.2   \\
    \rowcolor{ReactColor}
    SWE-Mirror-LM-32B~\cite{wang2025swe} & 32B & OpenHands & 52.2   \\
    \rowcolor{ReactColor}
    FrogBoss-32B~\cite{sonwane2025bugpilot} & 32B & OpenHands & 54.6   \\
    \rowcolor{ReactColor}
    Seed-OSS-36B~\cite{seed2025seed-oss}   & 36B & OpenHands & 55.2 \\
    \rowcolor{ReactColor}
    Llama3-SWE-RL-70B~\cite{wei2025swe} & 70B & OpenHands & 41.0   \\
    \rowcolor{ReactColor}
    Lingma-SWE-GPT-72B~\cite{ma2024lingma} & 72B & SWE-SynInfer & 28.8   \\
    \rowcolor{ReactColor}
    SWE-Fixer-72B~\cite{xie2025swe} & 72B & SWE-Fixer & 32.8   \\
    \rowcolor{ReactColor}
    GLM-4.5-Air  & 12/106B & OpenHands & 57.6   \\
    \rowcolor{ReactColor}
    Qwen3-235B-A22B~\cite{yang2025qwen3}  & 22/235B & OpenHands & 34.4   \\
    \rowcolor{ReactColor}
    Qwen3-Coder-480B-A35B~\cite{yang2025qwen3}  & 35/480B & OpenHands & 69.6   \\
    \rowcolor{ReactColor}
    DeepSeek-V3.1~\cite{liu2024deepseek}  & 37/671B & OpenHands & 61.0   \\
    \rowcolor{ReactColor}
    DeepSeek-R1-0528~\cite{guo2025deepseek} & 37/671B & OpenHands & 45.6   \\
    \rowcolor{HeaderColor}\multicolumn{4}{c}{\textbf{Summary Agent}} \\
    \rowcolor{SummaryColor}
    ReAct Agent   & 32B & OpenHands & 49.8 \\
    \rowcolor{SummaryColor}
    Threshold-Compression Agent   & 32B & OpenHands & 53.8 \\
    \rowcolor{HeaderColor}\multicolumn{4}{c}{\textbf{Folding Agent (Ours)}}\\
    \rowcolor{FoldingColor}
    \ourmodel{}   & 32B & OpenHands & \textbf{57.6} \\
    \bottomrule
    \end{tabular}%
    }
    \caption{Performance comparison on SWE-Bench Verified (N=500). We report Pass@1 results for different agent systems under a unified evaluation setting, grouped by model scale and agent framework.}
    \label{tab:main}
    \vspace{-10pt}
\end{table*}
\section{Experiments}
\paragraph{Datasets.} We evaluate the proposed method on the SWE-Bench-Verified~\cite{hou2024large} subset. SWE-Bench is a benchmark designed to assess the ability of LLMs to solve real-world software engineering tasks collected from 12 real-world GitHub repositories.  
The Verified split is a high-quality subset of SWE-Bench consisting of 500 instances, which are manually curated to provide clearer problem descriptions and more reliable evaluation criteria. We report solved rate as the primary evaluation metric, defined as the proportion of instances that are successfully resolved.

\paragraph{Training Data Construction.}
For training, we collect a large number of instances from two open-source datasets, SWE-smith~\cite{yang2025swe} and SWE-ReBench~\cite{badertdinov2025swe}. We first employ \datamethod{} to automatically generate agent interaction trajectories and apply a rejection sampling strategy to filter high-quality samples. This process yields a curated set of 20k supervised fine-tuning instances, referred to as \instruct{}, which effectively enhance the model’s context-management capability.
In addition, following the data construction protocol of SWE-smith, we collect an additional 20k high-quality supervised fine-tuning instances, denoted as \baseinstruct{}, which do not involve context-management skills. These data are used to train baseline models, ensuring fair and comparable evaluation against the proposed method.

\paragraph{Agent Post-training.}
We adopt Qwen2.5-Coder-32B~\cite{hui2024qwen2} as the base model and perform post-training on the \instruct{} dataset to obtain the final model, \ourmodel{}. The model is trained for up to three epochs using the AdamW~\citep{adamw} optimizer with a weight decay of 0.01. We employ a cosine learning rate schedule with a warm-up ratio of 0.1 and a peak learning rate of $5\times10^{-5}$.
During inference, we use the OpenHands~\cite{wang2024openhands} framework, where the agent can invoke tools including \texttt{execute\_bash}, \texttt{str\_replace\_editor}, \texttt{submit}, and \texttt{context}. For all experiments, the temperature is fixed to 0.0. The model is trained with a context length of $65{,}536$ tokens; for the evaluation reported in~\autoref{tab:main}, we allow the agent to perform up to 500 interaction rounds.

\paragraph{Baselines.}
We compare the proposed method with the following baselines: (1) ReAct~\cite{yao2022react}: This baseline follows the ReAct framework and does not employ any explicit context management.  Once the context window is exhausted, the dialogue terminates early. (2) Threshold-Compression~\cite{openhands_context}: This agent applies context compression only when the context length exceeds a predefined threshold. Upon triggering, it follows the same compression scheme as \ourmethod{}: the system prompt and key user intent are preserved verbatim, together with the most recent $k$ interaction messages, while all remaining earlier messages are summarized into a compact representation. For all baselines, we use the same base model as \ourmodel{} and use the same summarizer backbone for any compression operation. SFT is performed on the \baseinstruct{} dataset, which consists of 20k instances without context-management capabilities.
Besides, we also compare our method with existing closed-source and open-source systems, such as GPT-5 and DeepSeek-R1.

\paragraph{Main Results.}
\autoref{tab:main} presents the main experimental results, demonstrating the effectiveness of \ourmethod{}. On the challenging SWE-Bench-Verified benchmark, \ourmodel{} achieves a 57.6\% solved rate, reaching state-of-the-art performance under the setting of agent post-training on a 32B model. Under the same fine-tuning data budget, \ourmodel{} significantly outperforms both the ReAct Agent and the Threshold-Compression Agent baselines. Moreover, its performance is comparable to that of substantially larger models, and in some settings even surpasses them, under the same agent framework. These results indicate that \ourmethod{} and \datamethod{} are particularly effective for long-horizon interactive software engineering tasks such as those in SWE-Bench.

\section{Further Analysis}
\begin{figure}
    \centering
    \includegraphics[width=0.5\textwidth]{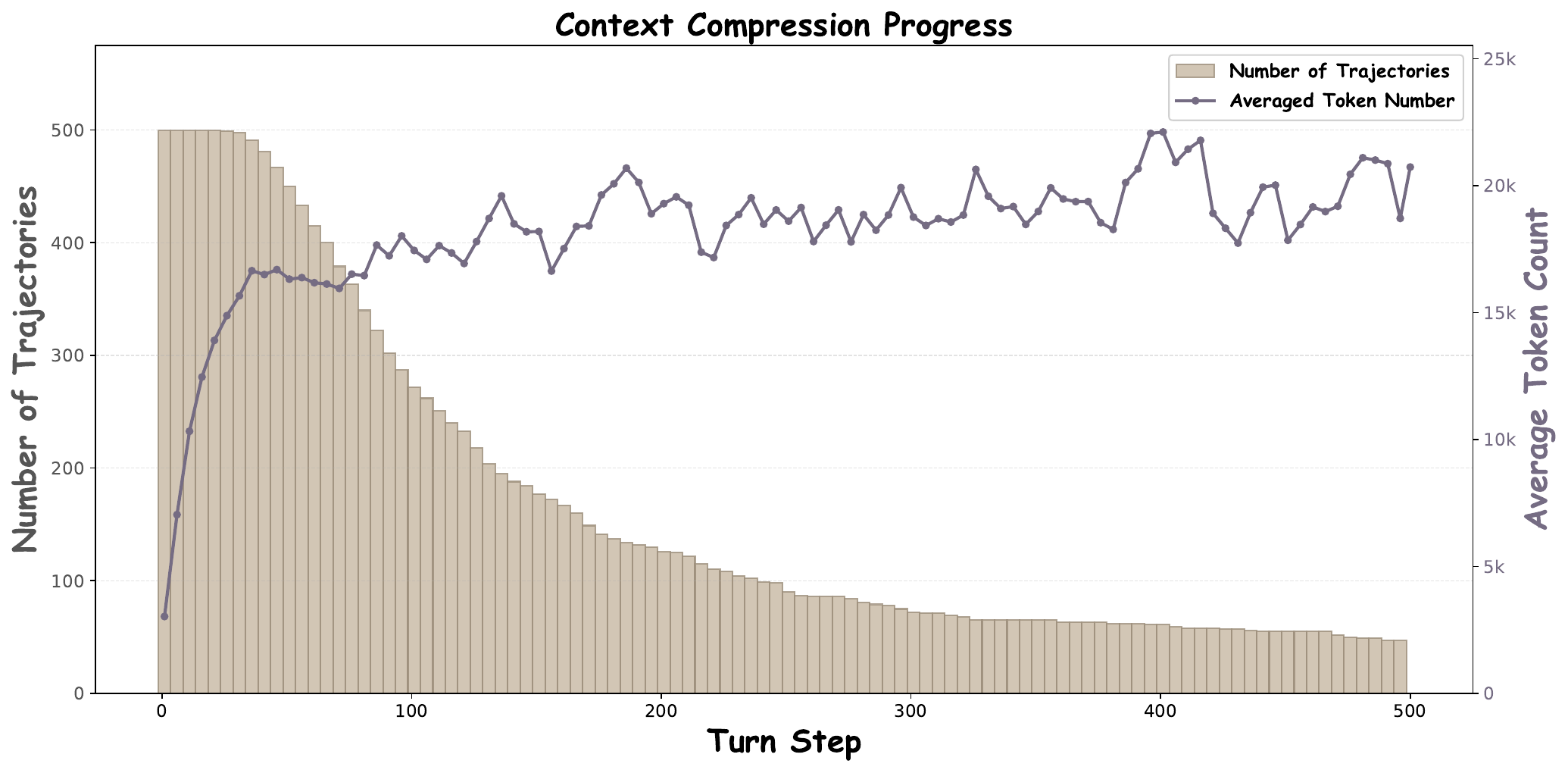}
    \caption{Context token usage and trajectory survival of \ourmethod{} over interaction rounds on SWE-Bench-Verified.}
    \label{fig:context_growth}
    \vspace{-10pt}
\end{figure}

\paragraph{Token Usage Analysis}
To evaluate the context management capability of \ourmethod{}, we analyze 500 interaction trajectories from SWE-Bench. Specifically, we report the number of surviving trajectories at each interaction round (\(|\mathcal{T}(t)|\)) and the average number of context tokens over the same set of trajectories at that round (\(A(t)\)).
The average context token count \(A(t)\) is formally defined as follows:
\begin{equation}
A(t) = \frac{1}{|\mathcal{T}(t)|} \sum_{j \in \mathcal{T}(t)} \text{TokenCount}\!\left(C^{(j)}_{t}\right)
\end{equation}
where \(\mathcal{T}(t)\) denotes the set of surviving trajectories whose lengths exceed \(t\) interaction rounds, and \(C^{(j)}_{t}\) represents the maintained context of trajectory \(j\) at round \(t\).
In~\autoref{fig:context_growth}, \ourmethod{} maintains a highly compact context. As the interaction progresses, the average token count quickly stabilizes after approximately 100 rounds and remains below 32k tokens, without exhibiting continuous growth over time, indicating the effectiveness of \ourmethod{} in preventing context explosion.
Moreover, the trajectory survival curve indicates that, in our experiments, more than 40\% of tasks remain interactive after 100 rounds. This observation further suggests that \ourmethod{} equips the model with stable and extensible long-horizon interaction capabilities, highlighting its strong potential for addressing highly complex and long-running software engineering tasks.

\begin{figure}[h]
	\centering
	\subfigure[Performance of varying interaction budgets.] {\includegraphics[width=0.4\textwidth]{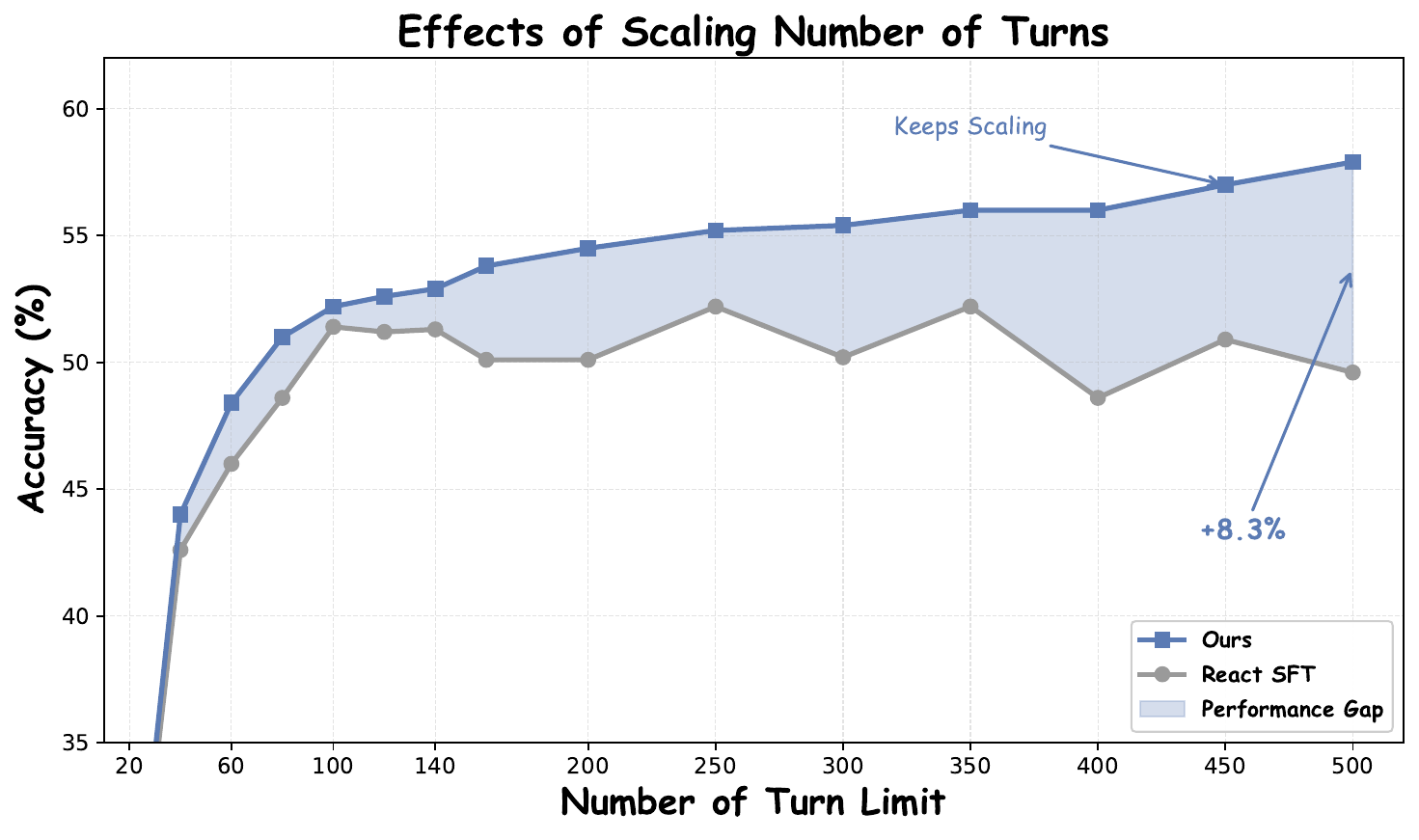}}
	\subfigure[Context token usage comparison.] {\includegraphics[width=0.4\textwidth]{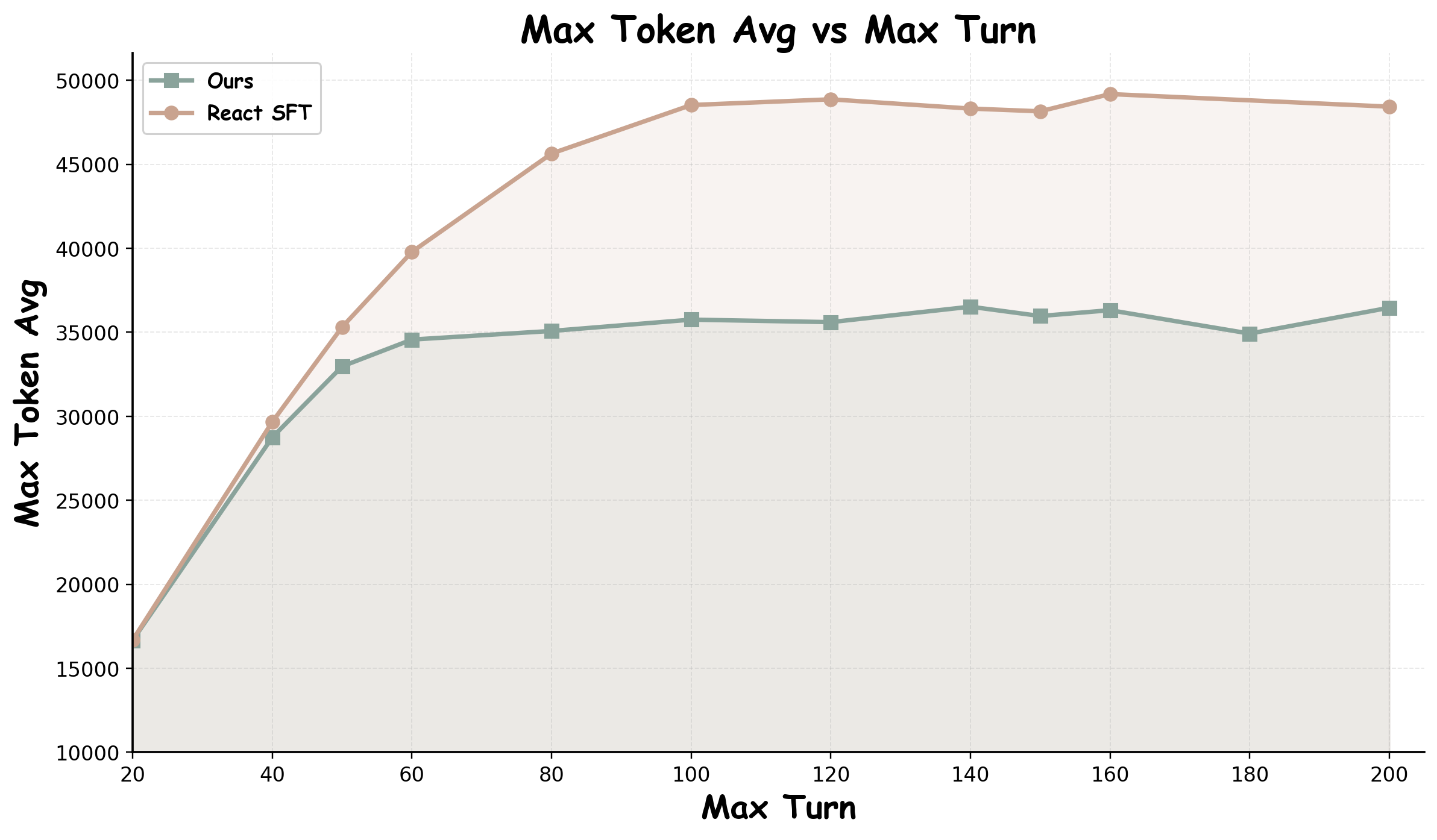}}
	\caption{Comparison of scalability and efficiency between \ourmethod{} and ReAct on SWE-Bench. (a) \ourmethod{} exhibits an upward trend in performance as the interaction budget increases to 500 rounds, whereas ReAct saturates and degrades after 60 rounds. (b) \ourmethod{} maintains stable context usage (35k tokens) via condensation, while ReAct rapidly exhausts the context window.}
	\label{fig:step_score_token}
    \vspace{-15pt}
\end{figure}

\paragraph{Context Comparison between \ourmethod{} and ReAct.}
\autoref{fig:step_score_token} evaluates \ourmethod{} and ReAct on SWE-Bench under identical maximum interaction round budgets. Two methods exhibit different behaviors across varying interaction budgets.
Across all comparable interaction budgets, the model equipped with \ourmethod{} consistently outperforms the SFT-based ReAct baseline, indicating more efficient utilization of historical information under the same reasoning budget. As the interaction budget increases, ReAct performance saturates after around 60 rounds and subsequently degrades, primarily due to its append-only context strategy: once the context window is filled, additional interactions fail to introduce effective information and instead hinder further reasoning.
In contrast, \ourmethod{} continues to improve steadily with increasing interaction budgets, maintaining a clear upward trend even at 500 rounds, suggesting that \ourmethod{} can continuously integrate salient information within a bounded context budget while compressing redundant history. In~\autoref{fig:step_score_token}, the context token usage of \ourmethod{} remains stable at approximately 35k tokens, whereas the ReAct baseline rapidly exhausts the available context window.

\begin{figure}
    \centering
    \includegraphics[width=0.5\textwidth]{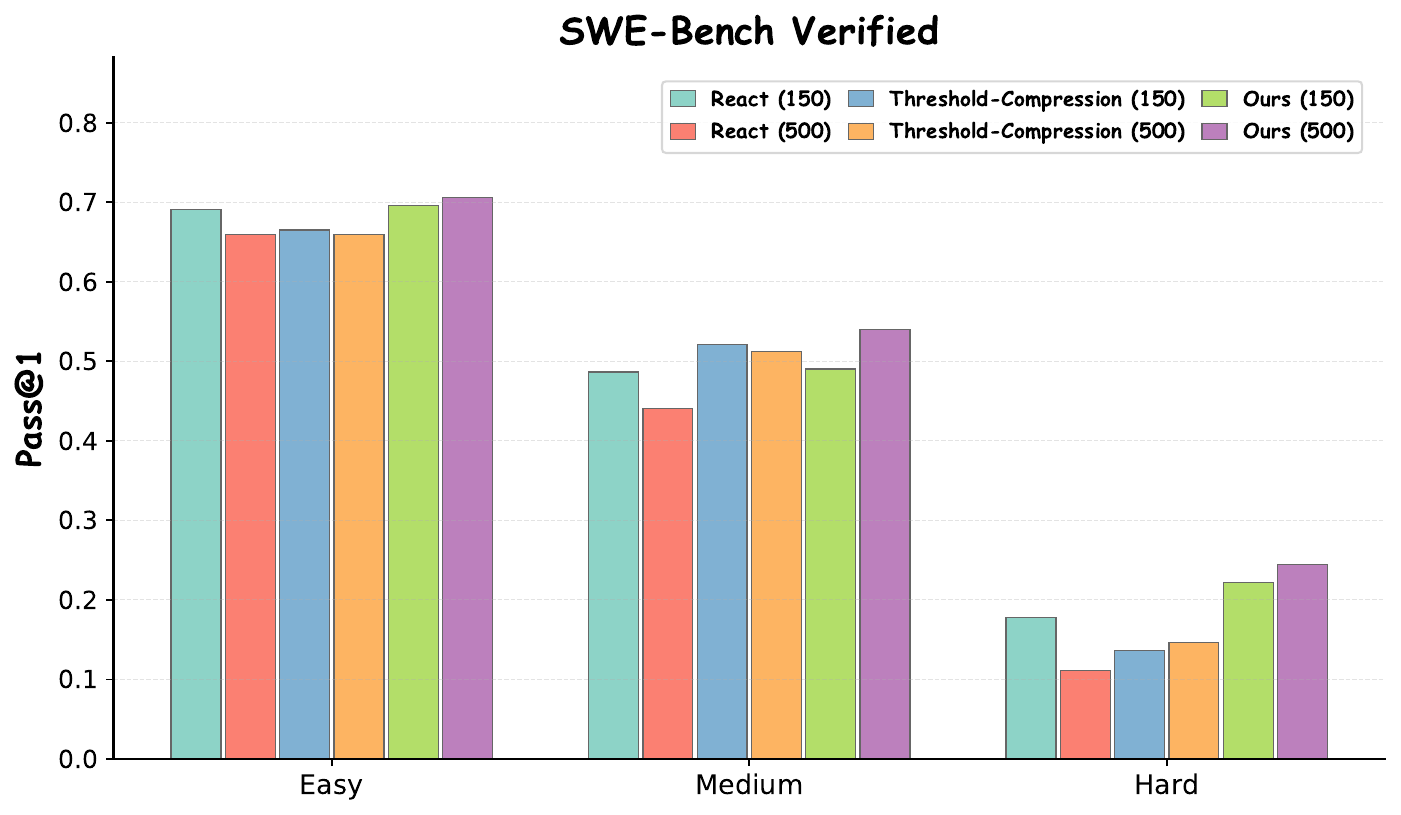}
    \caption{Performance classified by task difficulty on SWE-Bench Verified. \ourmethod{} outperforms baselines, with notably larger gains on medium and hard tasks that demand complex, long-horizon reasoning. }
    \vspace{-10pt}
    \label{fig:difficulty_scores}
\end{figure}

\paragraph{Performance by Task Difficulty.}
We partition SWE-Bench into difficulty levels based on the original dataset’s reported human solution time. Specifically, instances are categorized as easy ($\leq$15 minutes, 194 instances), medium (15 minutes--1 hour, 261 instances), and hard ($\geq$1 hour, 45 instances).
\autoref{fig:difficulty_scores} presents agent performance stratified by task difficulty, comparing the scores obtained under different context management strategies, showing that \ourmethod{} delivers stable and consistent performance improvements across easy, medium, and hard instances.
The performance gains are substantially larger on the medium and hard subsets than on the easy subset. This observation suggests that when tasks require more complex reasoning processes and longer-horizon context maintenance, the training signal introduced by \instruct{} becomes more effective. These findings further highlight the advantages of \datamethod{} and \ourmethod{} in addressing difficult tasks.
\begin{table}[t]
\centering
\label{tab:steps-tokens}
\resizebox{0.4\textwidth}{!}{%
\begin{tabular}{lccc}
\toprule
Method & Max Steps & Tokens & Pass (\%) \\
\midrule
ReAct & 150 & 1.96M & 53.2 \\
ReAct & 500 & 2.54M & 48.8 \\
\midrule
Threshold-Compression & 150 & 2.49M & 54.2 \\
Threshold-Compression & 500 & 5.18M & 53.8 \\
\midrule
\ourmethod{} (Base SFT) & 150 & 1.95M & 53.0 \\
\ourmethod{} (Base SFT) & 500 & 5.07M & 55.0 \\
\midrule
\textbf{\ourmethod{}} & 150 & \textbf{1.89M} & 54.8 \\
\textbf{\ourmethod{}} & 500 & 2.75M & \textbf{57.8} \\
\bottomrule
\end{tabular}
}
\caption{Performance and token usage under different maximum interaction steps on SWE-Bench-Verified.}
\label{tab:steps-tokens}
\vspace{-10pt}
\end{table}

\paragraph{Effect of Interaction Budget and Token Efficiency.}
\autoref{tab:steps-tokens} summarizes the performance and token usage of different methods under varying maximum interaction budgets. When a larger number of interactions is allowed (500 steps), \ourmethod{} achieves the highest pass rate among all methods, indicating stronger long-horizon reasoning capability.
Under a smaller interaction budget (150 steps), \ourmethod{} both maintains competitive performance and uses the fewest tokens across all methods. This favorable performance-efficiency trade-off is primarily attributed to the proposed context management mechanism, which prioritizes salient information while compressing redundant history, enabling efficient reasoning within a constrained context budget.
The comparison between \ourmethod{} and its Base SFT variant isolates the effect of \datamethod{}. Results show that \ourmodel{} consistently outperforms Base SFT model, especially under larger interaction budgets, confirming the importance of \datamethod{}.

\section{Related Work}
\label{sec:related_work}

\paragraph{Code Agents.}
As LLMs plateau on standalone code generation, recent work shifts toward agentic systems for real-world software engineering, typically evaluated on SWE-bench and Multi-SWE-bench~\cite{jimenez2023swe,zan2025multi}. Existing approaches improve performance either by enhancing agent designs with interactive tools and test-time scaling~\cite{wang2024openhands,yang2024swe,jain2025r2e,lin2025se,gao2025trae}, or by strengthening model-level agentic capability through large executable environments, synthetic supervision, and reinforcement learning~\cite{pan2024training,badertdinov2025swe,guo2025swe,yang2025swe,wang2025swe,sonwane2025bugpilot,wei2025swe,SWESwiss2025,deepswe2025}.

\paragraph{Context Management.}
To support long-horizon decision making, prior work explores context compression and memory mechanisms, including saliency-based filtering, hierarchical summarization, and multi-level memory architectures~\cite{li2023unlocking,jiang2023llmlingua,pan2024llmlingua,ye2025agentfold,sun2025scaling,packer2023memgpt,wang2023scm,hu2025hiagent,xiao2024infllm}. However, most rely on static compression or fixed memory policies, whereas our Tool Condensor enables dynamic, execution-driven context management that actively preserves decision-critical information over extended horizons.

\section{Conclusion}
In this work, we propose \ourmethod{}, a context management paradigm that treats context maintenance as a first-class, toolized capability in long-horizon agents. By integrating context management into the decision process of agent, \ourmethod{} enables proactive and structured condensation of interaction history, overcoming the limitations of append-only contexts and passive compression.
We further introduce a trajectory-level supervision framework with an offline retrofitting pipeline to inject context-management actions into full interaction trajectories. Experiments on SWE-Bench demonstrate that \ourmethod{} consistently outperforms ReAct agents and static compression baselines, while maintaining stable context usage and scalability under extended interaction budgets, underscoring the importance of modeling context evolution as an active and learnable component of agent behavior.



\newpage
\newpage
\bibliography{custom}

\newpage
\appendix

\section{Related Work}
\paragraph{Code Agents.} 
As LLMs approach saturation on traditional code generation tasks, research has shifted toward their agentic capabilities in real-world codebases, typically evaluated on SWE-bench and Multi-SWE-bench~\cite{jimenez2023swe,zan2025multi}. Existing efforts fall into two main directions.
The first focuses on agent design. OpenHands and SWE-Agent~\cite{wang2024openhands,yang2024swe} equip LLMs with interfaces such as editors and shells, enabling iterative file edits and command execution. Building on this setup, R2E-Gym, SE-Agent, and TraeAgent~\cite{jain2025r2e,lin2025se,gao2025trae} pursue test-time scaling by increasing sampling and decision steps to approximate upper-bound performance.
The second direction seeks to improve model-level agentic capability. SWE-Gym, SWE-rebench, and SWE-Factory~\cite{pan2024training,badertdinov2025swe,guo2025swe} build large executable environments for agent training, while SWE-Smith, SWE-Mirror, and BugPilot~\cite{yang2025swe,wang2025swe,sonwane2025bugpilot} generate synthetic tasks and trajectories for supervised fine-tuning. Reinforcement learning has also been explored: SWE-RL~\cite{wei2025swe} uses patch similarity as rewards, and SWE-Swiss and DeepSWE~\cite{SWESwiss2025,deepswe2025} investigate execution-based rewards as a promising alternative.

\paragraph{Context Management.} 
As agent technologies become crucial for long-horizon tasks, recent work aims to help LLMs sustain stable decision-making in complex environments. One line develops efficient context compression and memory management: methods like Selective Context, AgentFold, Context-folding, and LLMLingua~\cite{li2023unlocking,jiang2023llmlingua,pan2024llmlingua,ye2025agentfold,sun2025scaling} shorten inputs via salient-information filtering or hierarchical summaries, while AgentDiet and ACON~\cite{xiao2025improving,kang2025acon} enhance long-trajectory fidelity through reflection, contrastive learning, or latent-space compression. Another direction builds multi-level memory systems—MemGPT, SCM, HIAGENT, and InfLLM~\cite{packer2023memgpt,wang2023scm,hu2025hiagent,xiao2024infllm}—coordinating short- and long-term memory with retrieval modules for consistent extended reasoning. Multi-agent studies explore role-aware routing, shared memory, and divide-and-conquer frameworks to mitigate context explosion~\cite{liu2025rcr,tablan2025smarter,huan2025scaling}. Yet, these approaches rely on static compression, heuristic retrieval, or fixed memory, limiting adaptability and long-term coherence. In contrast, Tool Condensor enables dynamic, execution-driven context management, flexibly removing redundancy while preserving critical information—essential for complex long-horizon tasks.

\newpage
\end{document}